\documentclass[letterpaper, 10 pt, conference]{ieeeconf}
\IEEEoverridecommandlockouts
\usepackage{times}
\usepackage{cite}
\usepackage{graphicx}
\usepackage{hyperref}
\usepackage{float}
\usepackage{amsmath,amssymb}
\usepackage{booktabs}
\usepackage{multirow}
\usepackage{array}
\usepackage{soul}
\graphicspath{{./figures/}}

\title{\LARGE \bf
In-Situ Reconstruction of the International Space Station Using 3D Gaussian Splatting and Astrobee
}

\author{
Hudson Kim$^{1}$,
Ryan Soussan$^{2}$,
Brian Coltin$^{2}$,
Jordan Kam$^{3}$
\thanks{Jordan Kam is funded by NASA's Graduate Fellowship Program through the National GEM Consortium at NASA Ames Research Center.}
\thanks{$^{1}$ Department of Electrical Engineering and Computer Sciences, University of California, Berkeley, Berkeley, CA 94720, USA.}
\thanks{$^{2}$ Intelligent Robotics Group (KBR Inc.), NASA Ames Research Center, Moffett Field, CA 94035, USA.}
\thanks{$^{3}$ Graduate Aerospace Laboratories, California Institute of Technology, Pasadena, CA, 91125, USA.}
}

\begin{document}

\maketitle
\thispagestyle{empty}
\pagestyle{empty}

\begin{abstract}
This article presents a novel 3D reconstruction and mapping of the interior of the International Space Station (ISS) using 3D Gaussian Splatting (3DGS). Using existing grayscale images from the Astrobee free-flying robot dataset, we construct a full 3D splat of the ISS' Kibō or Japanese Experiment Module (JEM). 3DGS has in recent years shown promise in providing novel view synthesis of scenes captured from many images or videos, this article applies this approach to human spaceflight systems. We compare our 3DGS architecture to existing methods such as Nerfacto and TensoRF and show that reconstruction improves the state-of-the-art in both scene quality and rendering speed. We show that with as little as 500 in-situ images, a high-fidelity map can be constructed using Astrobee's Navigation Camera (NavCam) during free-flight in the JEM. These reconstructions could enable free-flyers to rapidly create and update interior maps for intra-vehicular habitats like the ISS.
\end{abstract}


\section{INTRODUCTION}

The interior of the International Space Station (ISS) changes constantly as cargo is transported \cite{morton2025deformable}, payloads are changed, and equipment is used by the crew. However, the current state-of-the-art 3D representation of ISS is a static computer aided design (CAD) model that is unable to capture these dynamic changes \cite{santos2024unsupervised}. This both limits current interior mapping capabilities, and makes decision-making difficult for ground-based mission controllers where having up-to-date interior representations of ISS could lead to more robust crew procedure synthesis. The Astrobee free-flying robots were engineered towards filling this gap \cite{smith2016astrobee}. Since 2019, Astrobee has been performing inspection, manipulation \cite{kam2025towards, chen2022testing}, and research tasks. Aimed to use robotic systems to reduce crew time on mundane tasks, Astrobee could routinely inspect ISS during downtime. 

Prior work has used Astrobee to gather large amounts of grayscale image data from its front-facing navigation camera (NavCam). This work has led to a large open-source dataset and benchmark for localization in the Japanese Experiment Module (JEM) all using Astrobee's NavCam \cite{kang2024astrobee}. Studies show that from this in-situ data, one can generate image feature maps for localization and denser 3D maps with higher-resolution imaging \cite{smith2023mapping}. While these results show great promise for interior mapping in-situ, these approaches have struggled in producing hyper photorealistic representations that can capture changes in the ISS without large amounts of images. Recent advances in neural and differentiable scene representations provide an alternative to existing work. Neural Radiance Fields (NeRF) \cite{mildenhall2020nerf} have demonstrated that ordinary posed images are sufficient to build photorealistic models of a scene. Following work \cite{muller2022instant} has led to the development of 3D Gaussian Splatting (3DGS) \cite{kerbl2023gaussian}, which replace the neural network part of NeRFs with explicit Gaussians. This has enabled faster training and real-time rendering. Terrestrial robots have already begun to adopt 3DGS for dense mapping and Simultaneous Localization and Mapping (SLAM) \cite{keetha2024splatam,matsuki2024gaussian} and have shown great success. State-of-the-art mapping approaches for ISS have relied on more classical techniques such as photogrammetric meshes \cite{smith2021isaac}, change detection \cite{dinkel2024astrobeecd}, and visual landmarks \cite{coltin2016imgfeaturemap}.

\begin{figure}[t]
\centering
\includegraphics[width=\linewidth]{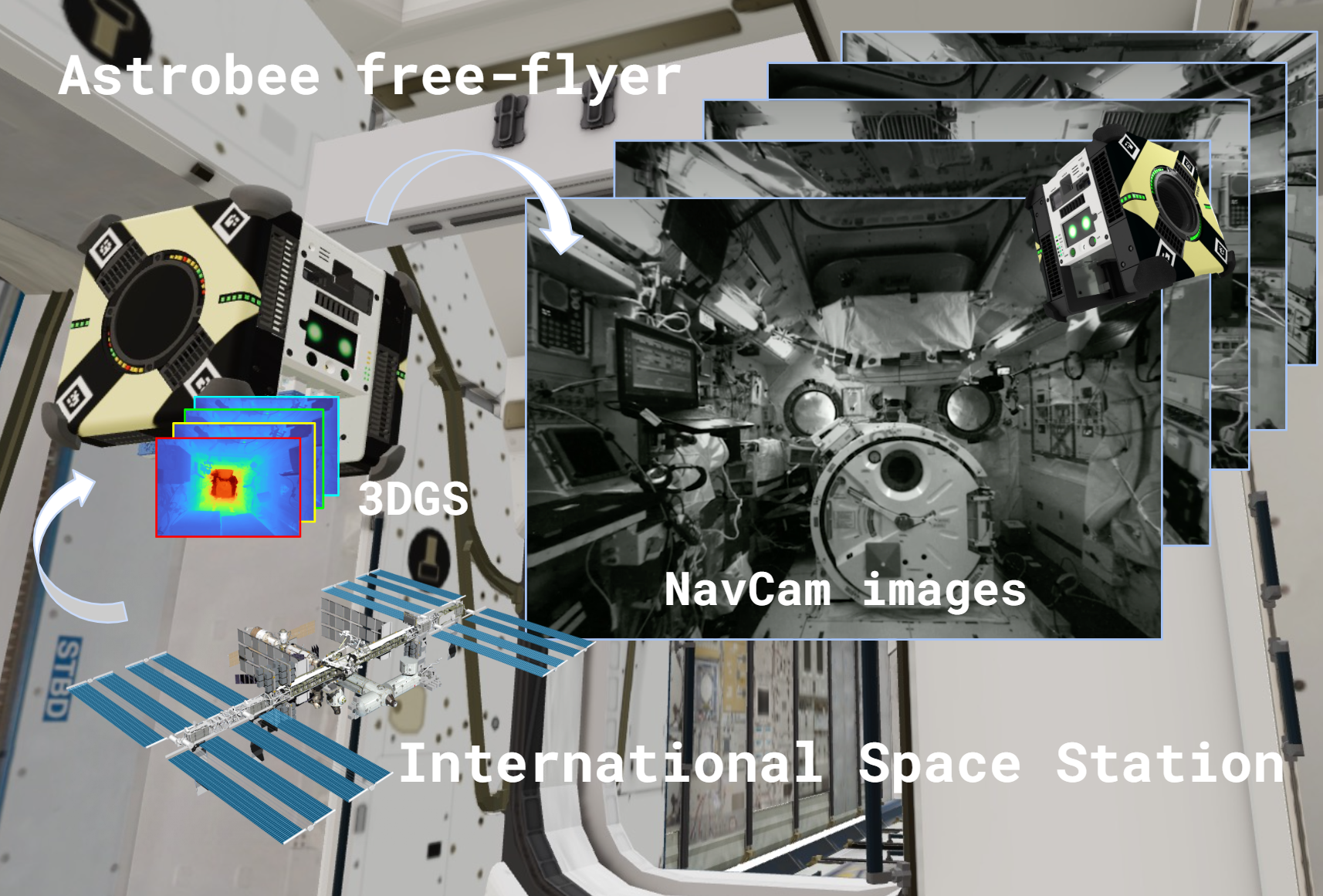}
\caption{Astrobee inside the ISS JEM Module with grayscale images and depth renders from in-situ NavCam images \cite{kang2024astrobee, kam2025towards}.}
\label{fig:astrobee}
\end{figure}

\begin{figure*}[t]
\centering
\includegraphics[width=\textwidth]{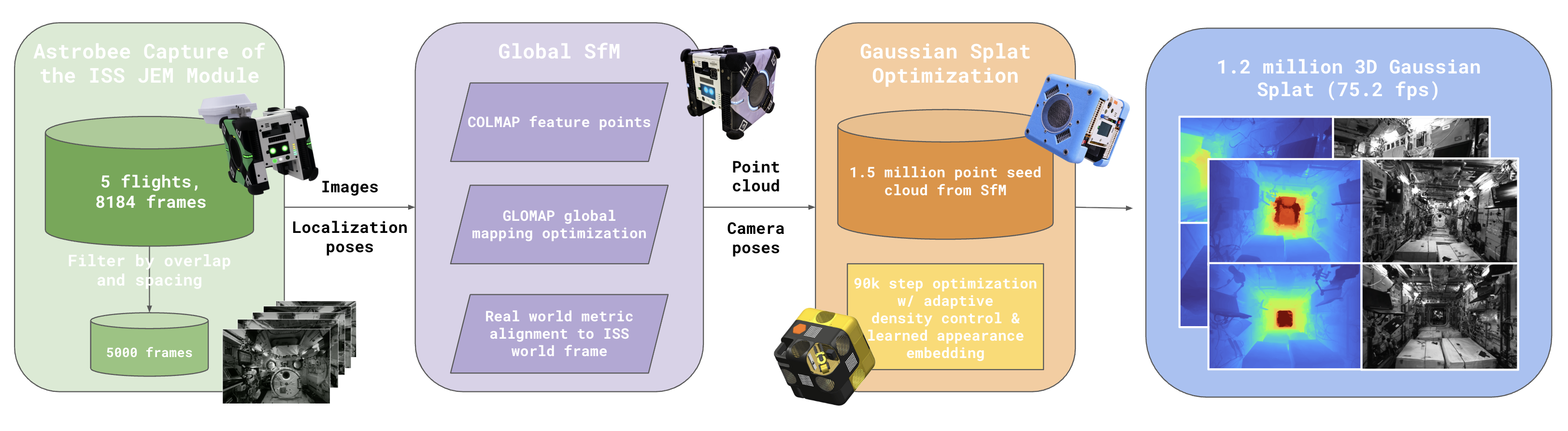}
\caption{Pipeline Diagram. Five Astrobee survey flights are filtered to 5{,}000 frames. COLMAP takes these images and identifies features. GLOMAP uses these features to simultaneously recover cameras poses and a 3D point cloud. The SfM data is then used as priors for the 3DGS map to be created while adaptive density control ensures that the correct number of Gaussians are used and a learned appearance embedding accounts for lighting differences.}
\label{fig:pipeline}
\end{figure*}

We present a pipeline for rapidly updatable interior maps of the ISS with 3DGS using as little as 500 images from the Astrobee free-flying robot dataset. We turn several Astrobee free-flights through the JEM into a 1.2 million 3D Gaussian Splat model able to render novel views in real time as shown in Fig.~\ref{fig:astrobee}. Our novel contributions include the following:

\begin{enumerate}
    \item To our knowledge, the first 3DGS reconstruction of the ISS.
    \item A scalable approach to take thousands of images from one or more trajectories and create a consistent 3DGS.
    \item Hyperparameter studies comparing how the number of optimizations and images impact the quality of a reconstruction as measured by PSNR.
\end{enumerate}

\section{METHODOLOGY}

Our pipeline shown in Fig.~\ref{fig:pipeline} has three stages, first we filter raw images from Astrobee into a smaller pool of useable images. From these, we then use classical Structure from Motion (SfM) techniques to recover camera poses for each frame and a point cloud of the surrounding environment. Finally, we use these point clouds and camera poses as priors to optimize a 3DGS model until its renders are able to match the ground truth photos. We selected five flights from April 2021 of the Astrobee free-flyer dataset that traverse the JEM and point Astrobee's NavCam forward, up, down, left, and right. To avoid redundant images, we sub-sample the original 8,184 raw images to 5,000 by only keeping frames in which Astrobee moved more than 1.5 cm or rotated more than 0.75°.

\subsection{Camera Poses from Structure from Motion}

Associated with each image in the dataset is a pose computed by Astrobee's visual localizer. While these poses are good for navigation, they lack the precision required to build a high-fidelity 3D reconstruction. We therefore recompute every pose from the images themselves using SfM, which detects distinctive feature points, matches them between frames, and solves for the camera poses and a sparse cloud of 3D points that best explain all the matches.  The most common SfM approach, incremental SfM, starts from a single image pair and adds one image at a time to a growing optimization problem. This works well for smaller datasets but scales quadratically in computation and relies on the images forming one well-connected set. Our input data however combines several runs through the JEM module, taken at different times and under different lighting conditions. This means that the same point in space taken from two flights can look considerably different and fail to be connected across frames. 

We work around this by connecting each flight's images through proposing matching candidate image pairs whenever Astrobee's visual localization poses indicate two images should visually overlap. For instance, if we have two images of a station tile from separate flights, we use Astrobee's pose associated with each frame to indicate these should be matched, even if the flights occurred in different lighting conditions or under other ISS changes. To avoid the fragile image-by-image growth of incremental SfM, we employ global SfM using Global Structure-from-Motion Revisited (GLOMAP) \cite{pan2024glomap} to jointly solve all rotations, positions, and points simultaneously. Compared to incremental SfM, GLOMAP averages out errors across the whole network while also being 10-100 times faster than COLMAP \cite{pan2024glomap}. As a result of our lighting-invariant SfM, we find a 10.25 dB improvement in PSNR and a reduction in SfM compute time of 9.7 minutes on a dataset of 549 images.

SfM variants, however, only recover the scene up to an unknown overall scale. In order to ensure real world alignment, we fit a similarity transform that aligns the SfM camera centers with the localization poses given from the Astrobee dataset. The fit sets the metric scale, and also measures how far Astrobee's visual localization poses and our own SfM poses disagree. We find a median disagreement of 6.7~cm between our SfM-based pipeline and Astrobee's localizer.

\begin{figure*}[t]
\centering
\includegraphics[width=\textwidth]{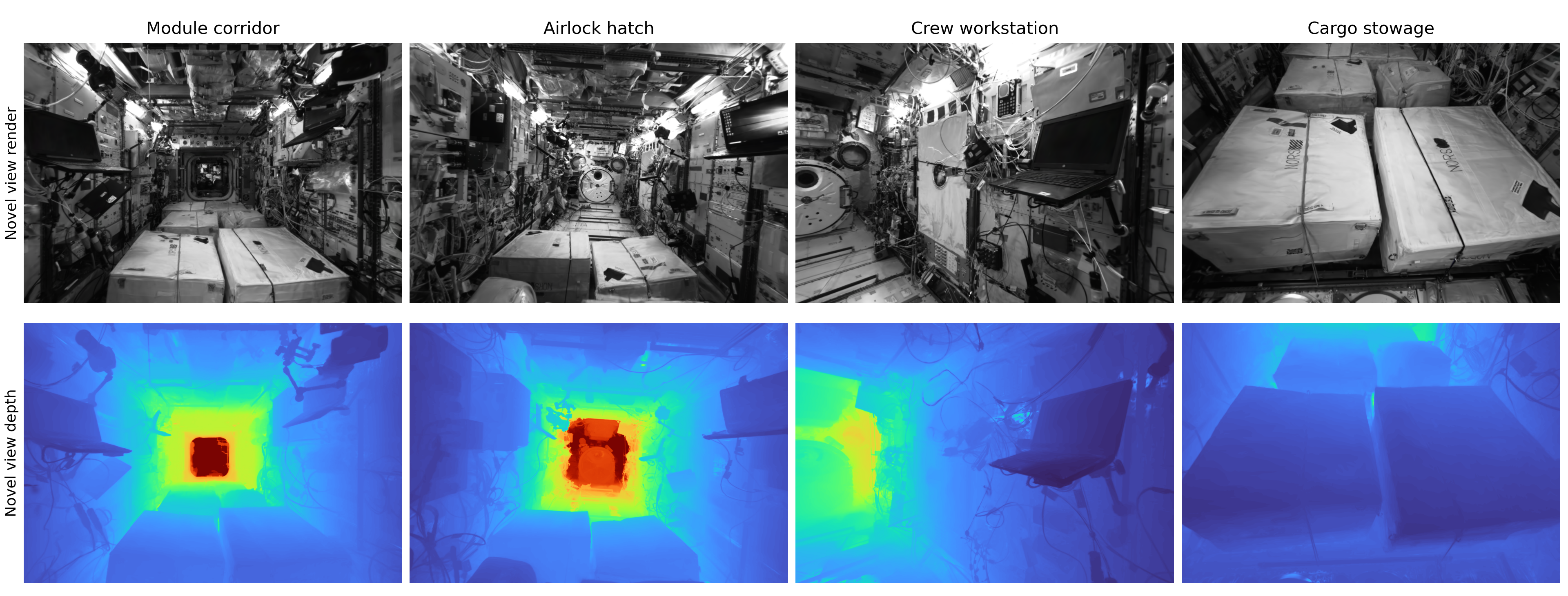}
\caption{Four novel views of the ISS Module corridor, Airlock hatch, Crew workstation, and Cargo stowage and their depth map from our 3DGS model.}
\label{fig:views}
\end{figure*}

\subsection{3D Gaussian Splatting Optimization}

Lastly, we build the reconstruction using the optimized camera poses and point cloud. Traditionally, 3DGS represents a scene as a large set of colored, semi-transparent, anisotropic, 3D Gaussians. Each Gaussian $k$ is defined as having a center $\mu_k$, an opacity $o_k \in [0,1]$, a color $c_k$ (represented with Spherical Harmonics for anisotropic coloring), and a covariance $\Sigma_k$ describing its ellipsoidal shape. The $\Sigma_k$ is factored as:
\begin{equation}
\Sigma_k = R_k S_k S_k^\top R_k^\top
\end{equation}
where $R_k$ defines the rotation and $S_k$ are the three axis lengths, ensuring the matrix remains positive semi-definite.

Using this, we place our initial Gaussians at the 1.5 million SfM points found previously to warm-start the optimization process. During optimization, we then begin to render viewpoints from the same poses as our ground truth image frames. To render a view from a desired camera pose, the Gaussians visible from that pose are projected into the image plane, sorted by depth, and alpha-blended front to back. The color at each pixel $\mathbf{p}$ of the render $C$ is defined as:
\begin{equation} C(\mathbf{p}) = \sum_{k} c_k\, \alpha_k \prod_{l<k} (1 - \alpha_l), \label{eq:composite} \end{equation} 
where the sum runs over the Gaussians along the ray through $\mathbf{p}$, ordered from nearest to farthest. Here, $c_k$ is Gaussian $k$'s view-dependent color evaluated from its spherical harmonics, and the blending weight $\alpha_k$ is its opacity $o_k$ attenuated by how far $\mathbf{p}$ falls from the center of its projected ellipse. Each pixel is therefore a weighted mixture of the Gaussians along its line of sight, with nearer, more opaque Gaussians occluding those behind them. 

Once a view is rendered, we optimize the scene. Because Eq.~\eqref{eq:composite} is fully differentiable, we optimize the reconstruction directly with gradient descent by rendering the scene from the pose of each training photo, comparing the render $C$ against the ground-truth image $I$, and updating the Gaussians by minimizing the loss, defined as:
\begin{equation} \mathcal{L} = (1-\lambda)\, \lVert C - I \rVert_1 + \lambda \big( 1 - \mathrm{SSIM}(C, I) \big), \label{eq:loss} \end{equation} 
Here, $\lambda = 0.2$, $C$ is the rendered image, and $I$ is the ground-truth image. The first term of the loss function penalizes average per-pixel error, while the Structural Similarity Index Measure (SSIM) term penalizes differences in local luminance, contrast, and structure. Additionally to better capture fine details in the scene, the number of Gaussians changes during training via adaptive density control. Every 100 steps, densification splits Gaussians whose image-space gradients stay large, implying the scene requires more detail at that point, while pruning deletes Gaussians that have gone nearly transparent or grown too big. We also consider  differences in exposure between flights, with each flight receiving a small learned appearance embedding, representing a global gain and bias, to be applied to its renders.

\section{RESULTS}

For evaluation, every eighth frame within the dataset is withheld from training (4{,}375 train / 625 eval views), and the scene is rendered at the evaluation image locations. These renders are then scored with PSNR, SSIM, and Learned Perceptual Image Patch Similarity (LPIPS) \cite{zhang2018lpips} at full resolution. Additionally, we consider the time it takes to run SfM and train the 3DGS reconstruction. Our method reaches 31.19~dB PSNR, 0.918 SSIM, and 0.208 LPIPS over the 625 ground-truth views, with 1.21M Gaussians rasterizing at 75.2~fps on an RTX~3090~Ti. Fig. ~\ref{fig:views} shows four novel views next to their rendered depth. 

\begin{table}[h]
\caption{Evaluation of the effect the number of images has on reconstruction quality and overall timing.}
\label{tab:img_scaling}
\centering
\footnotesize
\setlength{\tabcolsep}{3.5pt}
\begin{tabular}{@{}lrrrrrr@{}}
\toprule
\% of imgs & SfM & Train & Total & PSNR$\uparrow$ & SSIM$\uparrow$ & LPIPS$\downarrow$ \\
\midrule
100\% (4{,}375)  & 5.0~h & 47~m & 5.8~h & \textbf{31.19} & \textbf{0.918} & 0.208 \\
50\% (2{,}189)   & 1.7~h & 46~m & 2.5~h & 31.08 & 0.918 & 0.208 \\
25\% (1{,}095)   & 57~m  & 47~m & 1.7~h & 30.93 & 0.916 & 0.208 \\
12.5\% (549)     & 30~m  & 46~m & 1.3~h & 30.04 & 0.908 & 0.208 \\
6.25\% (277)     & 22~m  & 46~m & 1.1~h & 28.27 & 0.886 & 0.207 \\
3.125\% (140)    & 18~m  & 44~m & 1.0~h & 25.25 & 0.831 & 0.227 \\
\bottomrule
\end{tabular}
\end{table}

\begin{table}[h]
\caption{Evaluation of the effect the number of training iterations has on reconstruction quality and training time.}
\label{tab:iterations}
\centering
\footnotesize
\setlength{\tabcolsep}{5pt}
\begin{tabular}{@{}ccccc@{}}
\toprule
Iterations & PSNR$\uparrow$ & SSIM$\uparrow$ & LPIPS$\downarrow$ & Training Time\\
\midrule
15{,}000 & 25.29 & 0.836 & 0.305 & 7.4~m \\
30{,}000 & 20.61 & 0.716 & 0.410 & 16.2~m \\
45{,}000 & 29.10 & 0.902 & 0.228 & 24.3~m \\
60{,}000 & 30.09 & 0.911 & 0.217 & 32.0~m \\
75{,}000 & 30.89 & 0.916 & 0.211 & 39.0~m \\
90{,}000 & \textbf{31.19} & \textbf{0.918} & \textbf{0.208} & 45.6~m \\
\bottomrule
\end{tabular}
\end{table}

We also consider the effects hyper parameters have on the final reconstruction quality and total time. In Table~\ref{tab:img_scaling} and Table~\ref{tab:iterations}, we train the identical recipe with varying image counts and training iterations, testing to see its effect on reconstruction quality and training time. We find that greater than 500 images, the reconstruction quality does not scale linearly and leads to only marginal improvements. And greater than 60,000 training iterations, the reconstruction performance begins to plateau. 

Compared to existing reconstruction methods including Splatfacto 3DGS, Nerfacto, Instant NGP, and TensoRF, we find that our method improves on the current state-of-the-art, achieving higher scores on PSNR with 31.19 and render speed with 75.2 fps. Fig. 4 highlights these results. The improvement over the NeRF-based methods are likely from the use of Gaussians as an explicit scene representation, while the improvement over Splatfacto 3DGS can be attributed to our pose determination pipeline and learned appearance embedding attached to each flight to account for differences in lighting.

\begin{figure}[h]
    \centering
    \includegraphics[width=1\linewidth]{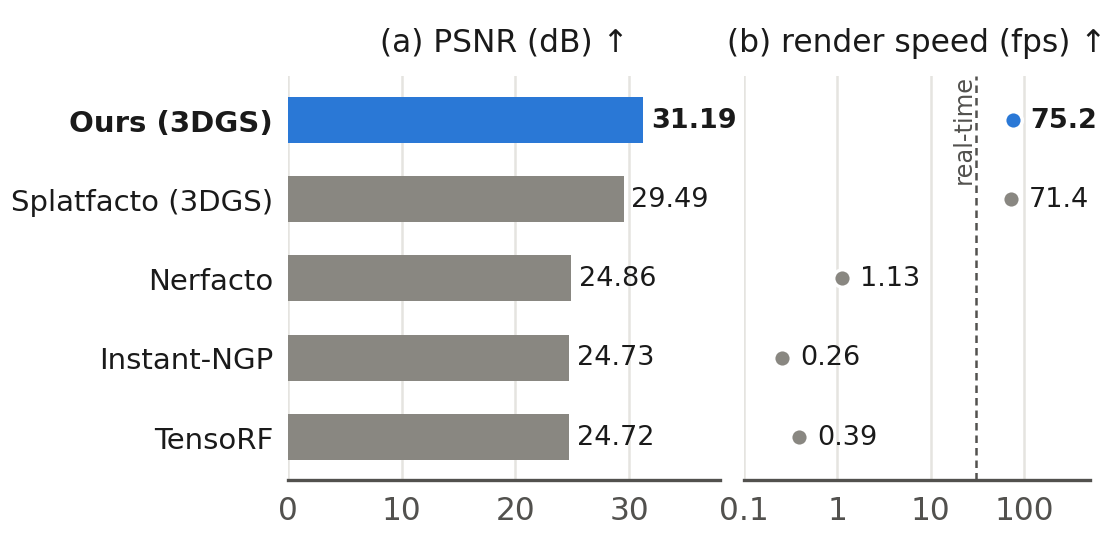}
    \caption{Comparison with Splatfacto, Nerfacto, Instant-NGP, and TensoRF.}
    \label{fig:placeholder}
\end{figure}

\section{CONCLUSION \& DISCUSSION}

We present the first 3DGS reconstruction using lighting-invariant SfM of the ISS, built from existing in-situ images of the ISS JEM. Our pipeline turns five disconnected survey free-flights into a single reconstruction that reaches 31.19~dB PSNR and renders at 75.2~fps on a consumer GPU, outperforming existing radiance field methods in quality and speed. A sufficient reconstruction takes 1.3 hours and $\sim$500 images, making routinely updated interior maps practical with existing perception hardware. These reconstructions can be used to assist in mission controller situational awareness by offering a photorealistic, freely explorable model for inspecting cargo stowage, panel configurations, and cable routing in-situ. Additionally, renders from arbitrary viewpoints can support crew procedure development and spaceflight training before a procedure is executed. Due to the model's scale, it can also serve as a simulation environment for robot planning and autonomy, enabling future free-flyer trajectories to be rehearsed against current ISS layouts.

There are several limitations with our current work. Firstly, our reconstruction inherits the limits of its input imagery. Astrobee's NavCam is grayscale and low-resolution, thus the model lacks precise RGB color and fine texture. Future work could incorporate Astrobee's color science camera to get a full RGB reconstruction with higher fidelity. Additionally, 3DGS also assume a static scene. The selected survey flights span multiple sessions between which the station may have changed. These inconsistencies currently show up as blurs in regions that moved. Finally, we have only reconstructed one module of the ISS. Validating the pipeline across the entire ISS remains for future work. 

\bibliography{references}

@article{morton2025deformable,
  title={Deformable cargo transport in microgravity with astrobee},
  author={Morton, Daniel and Antonova, Rika and Coltin, Brian and Pavone, Marco and Bohg, Jeannette},
  journal={arXiv preprint arXiv:2505.01630},
  year={2025}
}

@article{chen2022testing,
  title={Testing gecko-inspired adhesives with astrobee aboard the international space station: Readying the technology for space},
  author={Chen, Tony G and Cauligi, Abhishek and Suresh, Srinivasan A and Pavone, Marco and Cutkosky, Mark R},
  journal={IEEE Robotics \& Automation Magazine},
  volume={29},
  number={3},
  pages={24--33},
  year={2022},
  publisher={IEEE}
}

@article{kang2024astrobee,
  author  = {Kang, Suyoung and Soussan, Ryan and Lee, Daekyeong and Coltin, Brian and others},
  title   = {Astrobee {ISS} Free-Flyer Datasets for Space Intra-Vehicular Robot Navigation Research},
  journal = {IEEE Robot. Autom. Lett.},
  volume  = {9},
  number  = {4},
  pages   = {3307--3314},
  year    = {2024}
}

@inproceedings{santos2024unsupervised,
  title={Unsupervised change detection for space habitats using 3d point clouds},
  author={Santos, Jamie and Dinkel, Holly and Di, Julia and Borges, Paulo and Moreira, Marina and Alexandrov, Oleg and Coltin, Brian and Smith, Trey},
  booktitle={AIAA SCITECH 2024 Forum},
  pages={1960},
  year={2024}
}

@inproceedings{kam2025towards,
  title={Towards a microgravity sim-to-real training environment for robotic systems in low earth orbit},
  author={Kam, Jordan and Darrell, Trevor},
  booktitle={2025 Regional Student Conferences},
  pages={97394},
  year={2025}
}

@inproceedings{smith2023mapping,
  title={Mapping the ISS with the Autonomous Free-Flying Astrobee Robots},
  author={Smith, Trey and Bualat, Maria and Coltin, Brian and Akanni, Abiola and Alexandrov, Oleg and Benton, J and Hamilton, Kathryn and Moreira, Marina and Morris, Robert and Sharif, Khaled and others},
  booktitle={International Space Station Research and Development Conference (ISSRDC)},
  year={2023}
}

@inproceedings{mildenhall2020nerf,
  author    = {Mildenhall, Ben and Srinivasan, Pratul P. and Tancik, Matthew and Barron, Jonathan T. and Ramamoorthi, Ravi and Ng, Ren},
  title     = {{NeRF}: Representing Scenes as Neural Radiance Fields for View Synthesis},
  booktitle = {Proc. ECCV},
  year      = {2020}
}

@article{muller2022instant,
  author  = {M{\"u}ller, Thomas and Evans, Alex and Schied, Christoph and Keller, Alexander},
  title   = {Instant Neural Graphics Primitives with a Multiresolution Hash Encoding},
  journal = {ACM Transactions on Graphics},
  volume  = {41},
  number  = {4},
  pages   = {102:1--102:15},
  year    = {2022}
}

@article{kerbl2023gaussian,
  author  = {Kerbl, Bernhard and Kopanas, Georgios and Leimk{\"u}hler, Thomas and Drettakis, George},
  title   = {{3D} Gaussian Splatting for Real-Time Radiance Field Rendering},
  journal = {ACM Trans. Graph.},
  volume  = {42},
  number  = {4},
  year    = {2023}
}

@inproceedings{keetha2024splatam,
  author    = {Keetha, Nikhil and Karhade, Jay and Jatavallabhula, Krishna Murthy and Yang, Gengshan and Scherer, Sebastian and Ramanan, Deva and Luiten, Jonathon},
  title     = {{SplaTAM}: Splat, Track \& Map {3D} {Gaussians} for Dense {RGB-D} {SLAM}},
  booktitle = {IEEE/CVF Conference on Computer Vision and Pattern Recognition (CVPR)},
  year      = {2024}
}

@inproceedings{matsuki2024gaussian,
  author    = {Matsuki, Hidenobu and Murai, Riku and Kelly, Paul H. J. and Davison, Andrew J.},
  title     = {Gaussian Splatting {SLAM}},
  booktitle = {IEEE/CVF Conference on Computer Vision and Pattern Recognition (CVPR)},
  year      = {2024}
}

@inproceedings{smith2016astrobee,
  author    = {Smith, Trey and Barlow, Jonathan and Bualat, Maria and Fong, Terrence and Provencher, Christopher and Sanchez, Hugo and Smith, Ernest},
  title     = {Astrobee: A New Platform for Free-Flying Robotics on the {International Space Station}},
  booktitle = {International Symposium on Artificial Intelligence, Robotics, and Automation in Space (i-SAIRAS)},
  year      = {2016}
}

@inproceedings{smith2021isaac,
  author    = {Smith, Trey and Bualat, Maria and Akanni, Abiola and others},
  title     = {{ISAAC}: An Integrated System for Autonomous and Adaptive Caretaking},
  booktitle = {International Space Station Research and Development Conference (ISSRDC)},
  year      = {2021}
}

@article{dinkel2024astrobeecd,
  author  = {Dinkel, Holly and Di, Julia and Santos, Jamie and others},
  title   = {{AstrobeeCD}: Change Detection in Microgravity with Free-Flying Robots},
  journal = {Acta Astronautica},
  volume  = {223},
  pages   = {98--107},
  year    = {2024}
}

@inproceedings{pan2024glomap,
  author    = {Pan, Linfei and B{\'a}rath, D{\'a}niel and Pollefeys, Marc and Sch{\"o}nberger, Johannes L.},
  title     = {Global Structure-from-Motion Revisited},
  booktitle = {Proc. ECCV},
  year      = {2024}
}

@inproceedings{zhang2018lpips,
  author    = {Zhang, Richard and Isola, Phillip and Efros, Alexei A. and Shechtman, Eli and Wang, Oliver},
  title     = {The Unreasonable Effectiveness of Deep Features as a Perceptual Metric},
  booktitle = {Proc. CVPR},
  year      = {2018}
}

@inproceedings{coltin2016imgfeaturemap,
  author={Coltin, Brian and Fusco, Jesse and Moratto, Zack and Alexandrov, Oleg and Nakamura, Robert},
  booktitle={2016 IEEE/RSJ International Conference on Intelligent Robots and Systems (IROS)}, 
  title={Localization from visual landmarks on a free-flying robot}, 
  year={2016},
  volume={},
  number={},
  pages={4377-4382},
  doi={10.1109/IROS.2016.7759644}}
\bibliographystyle{IEEEtran}

\end{document}